\documentclass{article}

   \usepackage[nonatbib,final]{neurips_distshift_2022}

\usepackage[utf8]{inputenc} 
\usepackage[T1]{fontenc}    
\usepackage{hyperref}       
\usepackage{url}            
\usepackage{booktabs}       
\usepackage{amsfonts}       
\usepackage{nicefrac}       
\usepackage{microtype}      
\usepackage{xcolor}         

\usepackage{multirow}   
\usepackage{multicol}
\usepackage{amsthm,amsmath,amssymb} 
\usepackage{mathrsfs}
\usepackage{bm}
\usepackage{graphicx}
\usepackage{algorithm}
\usepackage{adjustbox,lipsum}
\usepackage{bigstrut}
\title{Toward domain generalized pruning by scoring out-of-distribution importance}

\author{%
  Rizhao Cai\\
   Nanyang Technological University \\
  \texttt{rzcai@ntu.edu.sg} \\
  \And
  Haoliang Li \\
  City University of Hong Kong \\
 \texttt{haoliang.li@cityu.edu.hk} \\
   \AND
  Alex Kot \\
   Nanyang Technological University \\
  \texttt{eackot@ntu.edu.sg} \\
}

\begin{document}

\maketitle

\begin{abstract}
    Filter pruning has been widely used for compressing convolutional neural networks to reduce computation costs during the deployment stage. Recent studies have shown that filter pruning techniques can achieve lossless compression of deep neural networks, reducing redundant filters (kernels) without sacrificing accuracy performance. However, the evaluation is done when the training and testing data are from similar environmental conditions (independent and identically distributed), and how the filter pruning techniques would affect the cross-domain generalization (out-of-distribution) performance is largely ignored. We conduct extensive empirical experiments and reveal that although the intra-domain performance could be maintained after filter pruning, the cross-domain performance will decay to a large extent. As scoring a filter's importance is one of the central problems for pruning, we design the importance scoring estimation by using the variance of domain-level risks to consider the pruning risk in the unseen distribution. As such, we can remain more domain generalized filters. The experiments show that under the same pruning ratio, our method can achieve significantly better cross-domain generalization performance than the baseline filter pruning method. For the first attempt, our work sheds light on the joint problem of domain generalization and filter pruning research.

\end{abstract}

\section{Introduction}
In the past decade, convolutional neural networks (CNNs) have been rapidly developed and widely applied for computer vision \cite{he2016deep, he2017mask}. For better precision performance, CNN models are gradually designed deeper and larger. However, this will lead to more model parameters and thus result in a higher burden of computation, making the model inefficient for real-time inference.

Recent research shows that CNNs models are usually overparameterized, and the overparameterized models have redundant parameters that can be pruned without performance decay \cite{LTH, taylorpruning, he2018amc}. Among different model compression techniques, filter pruning is an effective structured pruning method, which prunes redundant convolutional filters or kernels (hereafter, we use filters, kernels, or neurons alternatively). It has been shown that state-of-the-art filter pruning methods \cite{he2018amc, ding2021resrep, taylorpruning} can achieve a high compression ratio with little performance drop.  However, in the research community of neural network compression, the pruned network's performance is mainly evaluated based on the intra-domain setting, but the domain shift \cite{li2018domain, DomainBed} problem is largely ignored. Also, it has not been explored yet how the filter pruning would affect the performance when there exists domain shift between the training and testing data. We discuss the related works about generalization and pruning in Appendix~\ref{appendix-relatedwork}.

Motivated by the above discussion, we investigate how the filter pruning would affect the cross-domain generalization performance. We implement state-of-the-art filter pruning method \cite{taylorpruning}. Through experiments on the Domain Generalization (DG) benchmark, we find that the pruning method can preserve the intra-domain performance after pruning. However, the cross-domain testing shows significant performance decay when the distribution shift (domain shift) exists between the training and testing data. Therefore, we reveal that \textbf{pruning can degrade cross-domain generalization performance significantly}. This is a crucial problem urgent to be solved because the so-called ``lossless'' compressed models would work unexpectedly in the real world, where distribution shift problem usually appears.

To alleviate the problem of unexpected unseen-domain performance drops from pruning, we revisit existing pruning methods. In the common pruning-then-finetuning paradigm, \cite{abbasi2017structural, luo2017thinet, DBLP:conf/iclr/MolchanovTKAK17, ding2021resrep, taylorpruning, guo2020channel}, and a filter' importance score is calculated. Then, the filters with the lowest importance are masked or pruned. Thus, scoring a filter's importance is one of the central problems in pruning \cite{growing_reg}. Existing importance criteria are mainly based on magnitude \cite{ding2021resrep}, gradient information \cite{taylorpruning} or feature information \cite{guo2020channel}, which may work with the Independent and Identically Distributed (i.i.d) assumption, but cross-domain generalization is ignored which is also named the Out-of-Distribution (o.o.d) problem. Thus, directly pruning low-importance neurons will result in unexpected performance drops in the unseen target data domain.

\textcolor{black}{As the neurons' importance scoring is one of the central problems in pruning \cite{growing_reg}, we explore improving the importance estimation by taking into account the out-of-distribution risks. In this paper, we introduce the Importance of out-of-disribution Risk (IoR) to measure the filter importance, which is simple to implement by just including the gradient information of the variance of domain level risks to calculate the importance.}
In the experiments of the DG benchmarks PACS \cite{pacs_li2017deeper}, we evaluate both the intra-domain and cross-domain performance before and after pruning. Compared with the baseline pruning method \cite{taylorpruning}, our method with IoR can achieve better cross-domain generalization performance under the same pruning ratio.

\section{Preliminary}
A CNN model that has a number of $M$ filters can be parameterized as $\Theta = \{\theta_1, \theta_2, ... \theta_M \}$. Each filter $\theta_m$ is a convolutional kernel.  A heuristic solution is to select unimportant filters to be pruned and then finetune the pruned model \cite{taylorpruning}. Therefore, how to measure the importance of filters is one of the central problems \cite{growing_reg}.  When there $N$ multiple data domains $\mathbb{D}=\{D_1, D_2, ... D_N\}$ available, the importance can be expressed as 
\begin{equation}\label{eq-importance}
        I_m(\Theta) =  (\frac{1}{N}\sum_{i}^N\mathcal{R}_{i}(X,Y,\Theta) - \frac{1}{N}\sum_{i}^N\mathcal{R}_i(X,Y,\Theta|\theta_m=0))^2,
\end{equation}
where $\mathcal{R}_i$ denotes the empirical risk of domain $D_i$, and $\theta_m=0$ means that filter $\theta_m$ is pruned. Based on first-order taylor expansion \cite{taylorpruning}, Eq.\ref{eq-importance} can be approximated as 
\begin{equation} \label{eq-taylor}
        I_m(\Theta) = (\frac{1}{N}\sum_i^N\frac{\partial \mathcal{R}_i}{\partial \theta_m} \theta_m)^2 
    \end{equation}

As shown in Eq.~\ref{eq-importance}, the importance of filter $\theta_m$ is measured by evaluating how much risk is increased if $\theta_m$ is pruned ($\theta_m=0$). If the i.i.d assumption holds, the training data and testing data are drawn from the same distribution. As such, Eq.~\ref{eq-importance} can measure the importance of a filter w.r.t the testing data. However, distribution shift often exists in the real world, and testing data could be in a different distribution of training data. As such, directly using Eq.~\ref{eq-importance} for pruning could lead to unexpected performance drop in an unseen data distribution. To reduce the sacrifice of testing performance on the unseen data domain, we propose to select more generalized filters and pruned those less generalized filters by improving the importance scoring criterion. 
\section{Scoring out-of-distribution importance}
To improve the importance scoring, we consider the out-of-distribution risk to design the importance score. Inspired by risk extrapolation \cite{V-Rex}, we approximate the out-of-distribution risk $\mathcal{R}^o$ by
\begin{equation} \label{eq-ior}
    \mathcal{R}^o \approx Var\{{\mathcal{R}_1, \mathcal{R}_2,  ... \mathcal{R}_{N}}\},
\end{equation}
where $Var$ represents the variance of empirical risks from the source domains.

As such, the importance in Eq.~\ref{eq-taylor}  can be improved as $I_m^{IoR}$:
    \begin{equation} \label{eq-taylor-ior}
        I_m^{IoR}(\Theta) = (\frac{1}{N}\sum_i^N\frac{\partial \mathcal{R}_i}{\partial \theta_m} \theta_m)^2 + \alpha(\frac{\partial Var\{\mathcal{R}_1, \mathcal{R}_2,  ... \mathcal{R}_N\}}{\partial \theta_m} \theta_m)^2,
    \end{equation}
where $\alpha$ is a constant scaling factor and $\alpha=1$ in our experiments. 
Since the gradient $\frac{\partial \mathcal{R}_i}{\partial \theta_m}$ can be obtained during the training and pruning, there would not be much extra computation needed for calculating Eq.~\ref{eq-taylor}. 
During pruning, importance scores of all filters are obtained. The filters are ranked according to importance scores in ascending order. In each iteration, top-K remained filters are removed and then finetuned.

\section{Experiments}
\subsection{Implementation}
We follow the leave-one-domain-out protocol and the setting used in \cite{digit_zhou2020learning}. The setup of data of training, (intra-domain) validation, and (cross-domain) testing can be found in Appendix~\ref{appendix-protocol}.  To implement the baseline, we use the official implementation of Taylor Pruning  with the suggested hyper-parameters \cite{taylorpruning}. Detailed parameters setting can be found in Appendix~\ref{appendix-implementation}.  Our proposed method also uses the same hyperparameters. All experimental results are reported by averaging the results from five independent runs with different random seeds. 

\begin{table}[tbp]
  \centering
  \caption{Pruning ResNet18 and ResNet50 models on the PACS benchmark. ``Intra” means the intra-domain top-1 accuracy (\%), and “Cross” means the
cross-domain top-1 accuracy (\%)}
  \small
    \begin{tabular}{l|cc|cc|cc|cc}
    \hline
    \multicolumn{1}{c|} {Model} & \multicolumn{2}{c|}{Art} & \multicolumn{2}{c|}{Cartoon} & \multicolumn{2}{c|}{Photo} & \multicolumn{2}{c}{Sketch} \\
\cline{2-9}          & Intra & Cross & Intra & Cross & Intra & Cross & Intra & Cross \\
    \hline
    ResNet18 (before pruning) & 99.63 & 76.86 & 100   & 78.20 & 97.39 & 93.59 & 99.68 & 77.86 \\
    \hline
    ResNet18-50\% (baseline)& 99.28	&74.94&	99.97&	77.50&	95.82	&92.04&	99.27	&\textbf{77.41}\\
    ResNet18-50\% with IoR (ours) & 99.34 & \textbf{75.60} &99.90  & \textbf{78.79} & 96.80& \textbf{92.53} & 99.42 &75.98 \\
    \hline
    ResNet18-30\% (baseline) & 97.12 & \textbf{68.85} & 97.86 & \textbf{73.74} & 91.34 & 85.09 & 96.79 & 69.51 \\
    ResNet18-30\% with IoR (ours) & 95.94 & 68.37 & 97.59 & 72.31 & 92.98&\textbf{86.37} & 96.55 & \textbf{70.81} \\
    \hline
    \hline
    ResNet50  (before pruning) & 95.04 & 83.50 & 96.78 & 81.27 & 96.56 & 96.11 & 99.84 & 83.53 \\
    \hline
    ResNet50-50\% (baseline) & 99.13 & 82.84 & 99.82 & 80.30 & 98.49 & 93.68 & 100 & 79.04 \\
    ResNet50-50\% with IoR (ours) & 99.22 & \textbf{83.36}  & 98.36 &\textbf{80.56}  &  98.10  & \textbf{94.49}  & 99.97 & \textbf{79.74}\\
    \hline
    ResNet50-30\% (baseline) &  97.37&	74.53	&97.99	&76.79&	93.14&	86.86&	98.57	&74.87\\
    ResNet50-30\% with IoR (ours) & 97.55 & \textbf{75.76} &    98.12   & \textbf{77.11}  & 96.68 & \textbf{91.47} & 98.57 & \textbf{78.36}
    
 \\
    \hline
    \end{tabular}%
  \label{tab:pacs-baseline}%
\end{table}%

\begin{table}[htbp]
  \centering
  \small
    \caption{Pruning 50\% of the filters of the ResNet models pretrained by ERM, CORAL, and Mixup on the PACS benchmark. ``Intra'' means the intra-domain top-1 accuracy (\%), and ``Cross'' means the cross-domain top-1 accuracy (\%).}
    \begin{tabular}{c|c|cc|cc|cc|cc}
    \hline
    \multirow{2}[4]{*}{Pretrain Method} & \multirow{2}[4]{*}{Model} & \multicolumn{2}{c|}{Art} & \multicolumn{2}{c|}{Cartoon} & \multicolumn{2}{c|}{Photo} & \multicolumn{2}{c}{Sketch} \bigstrut\\
\cline{3-10}          &       & Intra & Cross & Intra & Cross & Intra & Cross & Intra & Cross\\
    \hline
    \multirow{3}[2]{*}{ERM} & \multicolumn{1}{l|}{ResNet50} & 95.04 & 83.50  & 96.78 & 81.27 & 96.56 & 96.11 & 99.84 & 83.53 \\
          & \multicolumn{1}{l|}{ResNet50-50\%} & 99.13 & 82.84 & 99.82 & 80.30  & 98.49 & 93.68 & 100   & 79.04 \\
          & $\Delta$     & 4.09  & -0.66 & 3.14  & -0.97 & 1.93  & -2.43 & 0.16  & -4.49 \\
    \hline
    \multirow{3}[2]{*}{CORAL} & \multicolumn{1}{l|}{ResNet50} & 96.15 & 87.06 & 98.45 & 83.96 & 99.29 & 96.47 & 97.57 & 81.65 \\
          & \multicolumn{1}{l|}{ResNet50-50\%} & 96.77 & 80.37 & 98.20  & 76.83 & 99.17 & 94.49 & 97.08 & 73.02 \\
          & $\Delta$     & 0.62  & -6.69 & -0.25 & -7.12 & -0.12 & -1.98 & -0.49 & -8.63 \\
    \hline
    \multirow{3}[2]{*}{Mixup} & \multicolumn{1}{l|}{ResNet50} & 96.52 & 87.55 & 94.47 & 82.25 & 98.7  & 97.13 & 99.03 & 80.63 \\
          & \multicolumn{1}{l|}{ResNet50-50\%} & 96.4  & 79.15 & 96.14 & 79.14 & 98.58 & 94.91 & 98.21 & 69.41 \\
          & $\Delta$    & -0.12 & -8.40  & 1.67  & -3.11 & -0.12 & -2.22 & -0.81 & -11.20 \\
    \hline
    \end{tabular}%
  \label{tab:dg}%
\end{table}%

\subsection{Experimental Results}
We present the results by asking three research questions.\\
 \textbf{Research Question Q1: Will filter pruning affect the domain generalization performance? }\\
 Following DomainBed\footnote{https://github.com/facebookresearch/DomainBed} \cite{DomainBed}, we first use the Empirical Risk Minimization (ERM) to pretrain the ResNet18 and ResNet50 models, which are often used as backbones in domain generalization benchmarks \cite{NCDG, wang2022domain}. Then, we use the baseline pruning method \cite{taylorpruning} to prune the pretrained ResNet18 and ResNet50 models. After pruning, 50\% or 30\% of filters remain, and the pruned models are denoted as ResNet18/50-50\% or ResNet18/50-30\%, respectively, as shown in Table~\ref{tab:pacs-baseline}. We evaluate the top-1 accuracy of intra-domain and cross-domain accuracy before and after pruning. 
 
Table~\ref{tab:pacs-baseline} shows the pruning results on the PACS benchmark. When we compare the intra/cross-domain accuracy before and after pruning, we find that 1) \textit{cross-domain generalization performance is not guaranteed even the intra-domain evaluation shows ``lossless'' compression results}. As shown in Table~\ref{tab:pacs-baseline},  when the pruning to 50\% or 30\%, the intra-domain accuracy could have insignificant decay and could even improve to some extent, which indicates the ``lossless'' pruning. However, we can see the pruned models' cross-domain generalization performance degrades significantly in different experiments (Art, Cartoon, Photo, Sketch) under different compression ratios. The most interesting experiment is the \textit{Photo} experiment of ResNet50. It appears that the original model has good generalization performance since the cross-domain accuracy is close to intra-domain accuracy before pruning (96\%). However, after pruning (50\% or 30\%), the compressed model's cross-domain accuracy results are far lower than intra-domain results. 

 \textbf{Research Question Q2: Can we reduce the sacrifice of the cross-domain performance from pruning? }\\
We explore this question by using our proposed importance scoring based on out-of-distribution risk. In Table~\ref{tab:pacs-baseline}, we compare the pruning results of the baseline and our method (with IoR). When comparing the cross-domain accuracy of the compression ratios 50\% and 30\% for ResNet18 and ResNet50, there are 16 groups to be compared. Our IoR can achieve better results in 13 out of 16 comparison pairs (see the bold \textbf{accuracy} results). As such, more cross-domain generalization performance can be saved from pruning by using our IoR. Thus, our work shows that designing a better importance criterion is a feasible way to save cross-domain generalization from pruning. In future work, we could design better criteria to further save the cross-domain generalization performance.

\textbf{Research Question Q3: Is a model pretrained by domain-generalization methods more than ERM robust after pruning? }
We also investigate if a model pretrained by using domain generalization methods can be more robust during the pruning. Similar to Q1, we follow DomainBed \cite{DomainBed} to use the domain generalization methods CORAL \cite{CORAL} and Mixup \cite{mixup} to pretrain models before filter pruning.
Table~\ref{tab:dg} compares the accuracy results of ResNet50 models pretrained by ERM, CORAL, and Mixup, before and after pruning 50\% of filters, on the PACS benchmarks. In the experiments of Art, Cartoon and Photo, CORAL and Mixup can help to achieve higher cross-domain accuracy results than ERM (before pruning). Also, for all experimental results of CORAL and Mixup in Table~\ref{tab:dg}, the intra-domain accuracy performance is well maintained, with a drop of 0.81\% at most. However, after pruning, we observe that the cross-domain accuracy results of CORAL and Mixup generally drop far more than that of the ERM. We conjecture the reason that in the pruning-then-finetuning paradigm, the finetuning is based on ERM, which has different optimization processes from the DG optimization methods. Therefore, when pruning the CORAL and Mixup pretrained models, finetuning the pruned models with ERM could make the optimization process inconsistent, leading to poorer performance.

\section{Conclusion}
In this work, we investigate how filter pruning would affect the cross-domain generalization when there exists domain shift between training and testing data. By conducting the experiments to answer the three research questions we raise, we reveal that: 1) \textit{ ``lossless" compression can lead to significant cross-domain performance drop, even with few decays of intra-domain accuracy}; 2) \textit{using domain generalization methods to pretrain a more generalized model (than ERM) could not help to reserve the drop accuracy from pruning}; 3) \textit{our IoR can alleviate the cross-domain accuracy drop from pruning by a better importance criterion}.
Our work shed the light on the joint problem of pruning and domain generalization. We wish our work can attract more research attention to the joint problem of pruning and domain generalization.

\bibliographystyle{plain}
\bibliography{main}

\appendix

\section{Appendix}
\subsection{Related works}\label{appendix-relatedwork}
With regards to pruning and generalization, the most relevant work is \cite{liebenwein2021lost}. While \cite{liebenwein2021lost} mainly focuses on the pruning and robustness against corruption. Our work mainly focuses on the pruning and domain generalization toward ``covariate shift".

Also, there are other papers about cross-domain compression \cite{sinno_pruning, feng2020admp},  aiming to utilize the source domain data to improve the pruning performance in the target domain. In \cite{sinno_pruning, feng2020admp}, the target domain data is available for training and pruning, which is similar to the \textbf{domain adaptation} setting \cite{tzeng2017adversarial}. However, our work is related to the \textbf{domain generalization} setting \cite{li2018domain, zhou2021domain}, where the target domain data is unseen (unavailable) during the pruning process.

\subsection{Evaluation protocol}\label{appendix-protocol}
We follow the leave-one-domain-out protocol and the setting used in \cite{digit_zhou2020learning}. The data in PACS of each domain is split into two parts according to training : validation = 9 : 1. In each experiment, one domain is left as the unseen target domain for testing, while the rest data of the other domains is used as source domains for training. For example, in the experiment of Photo of the PACS benchmark, the training parts of Art, Cartoon, and Sketch are combined for the model pretraining and pruning. Meanwhile, the validation parts of Art, Cartoon, and Sketch are combined to validate the top-1 accuracy. The pruned model with the best top-1 validation accuracy is selected for the testing, and the best validation accuracy is reported as the intra-domain performance. After that, we use all the data of the Photo domain as the unseen target domain to test the cross-domain performance in experiments of before or after pruning.

\subsection{Implementation details}\label{appendix-implementation}
As the baseline pruning method is equivalent to the Taylor pruning \cite{taylorpruning}, we use its official PyTorch implementation \footnote{https://github.com/NVlabs/Taylor\_pruning}  to conduct baseline pruning experiments. After each convolutional layer, we utilize the gate replacement to use the gate's importance to represent its filter's importance  \cite{taylorpruning} . Following the recommended settings \cite{taylorpruning}, we use batch size 64, learning rate 0.001, and epochs 100. The pruning interval is 30 mini-batches, during which a maximum of 100 filters would be pruned.  Also, we use the exponential moving average filter (momentum) with a coefficient 0.9 to calculate the importance scores over different mini-batches.  The experimental results are reported by averaging the results five independent runs with different random seeds.

\end{document}